\documentclass{article}
\usepackage{ijcai26}

\usepackage{times}
\usepackage{soul}
\usepackage{url}
\usepackage[hidelinks]{hyperref}
\usepackage[utf8]{inputenc}
\usepackage[small]{caption}
\usepackage{graphicx}
\usepackage{amsmath}
\usepackage{amsthm}
\usepackage{booktabs}
\usepackage{algorithm}
\usepackage{algorithmic}
\usepackage[switch]{lineno}

\usepackage{dsfont}
\usepackage{tikz}
\usepackage{todonotes}
\usepackage{adjustbox}
\usepackage{wrapfig}
\usepackage{rotating}
\usepackage{paralist}
\usetikzlibrary{shapes.geometric, arrows.meta, positioning}
\usetikzlibrary{decorations.markings}

\DeclareMathOperator*{\argmin}{arg\,min}

\title{Out-of-Distribution Detection Using Counterfactual Distance}

\author{
Maria Stoica
\and
Francesco Leofante\and
Alessio Lomuscio\\
\affiliations
Department of Computing, Imperial College London, UK\\
\emails
\{m.stoica22, f.leofante, a.lomuscio\}@imperial.ac.uk
}

\begin{document}

\maketitle

\begin{abstract}
Accurate and explainable out-of-distribution (OOD) detection is required to use machine learning systems safely.
Previous work has shown that feature distance to decision boundaries can be used to identify OOD data effectively.
In this paper, we build on this intuition and propose a post-hoc OOD detection method that, given an input, calculates the distance to decision boundaries by leveraging counterfactual explanations.
Since computing explanations can be expensive for large architectures, we also propose strategies to improve scalability by computing counterfactuals directly in embedding space.
Crucially, as the method employs counterfactual explanations, 
we can seamlessly use them to help interpret the results of our detector.
We show that our method is in line with the state of the art on CIFAR-10, achieving 93.50\% AUROC and 25.80\% FPR95.
Our method outperforms these methods on CIFAR-100 with 97.05\% AUROC and 13.79\% FPR95 and on ImageNet-200 with 92.55\% AUROC and 33.55\% FPR95 across four OOD datasets.
\end{abstract}

\begin{figure}[!ht]
    \centering
    \includegraphics[width=0.3\textwidth]{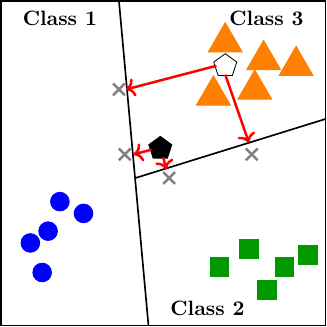}
\caption{We present an instantiation of our framework with a toy example of a three-class classification problem. In-distribution points are represented by the coloured circles, triangles, and squares. We then consider two new inputs: an OOD input (black pentagon) and an ID input (white pentagon). Assume both inputs are initially classified as class 3. In this example, we compute two counterfactuals (the grey crosses) for each input, one for class 1 and one for class 2, and compute the distances to these counterfactuals. We observe that the OOD input is closer to the decision boundaries and thus results in a smaller average counterfactual distance. Instead, the ID input is farther from the decision boundaries and thus will have a greater average counterfactual distance than the anomaly point. In this paper, we leverage this intuition to build an OOD detection method that uses counterfactual distances as OOD scores to identify anomalies.}
\label{fig:illustration}
\end{figure}

\section{Introduction}\label{sec:intro}

As deep neural networks (DNNs) are deployed in critical applications, safety research has emerged to ensure reliable AI systems. While DNN verification methods provide design-time safety assurances assuming static operating conditions~\cite{liu_algorithms_2021}, safely handling runtime distribution shifts remains a major concern. Out-of-distribution (OOD) detection addresses this by identifying data points that differ from the in-distribution (ID) training data~\cite{amodei_concrete_2016,hendrycks_baseline_2017}.
 
Several OOD detection methods exist. 
Their common aim is to differentiate between ID and OOD data using different techniques.  
One such category of detectors is \textit{post-hoc} detectors, which are able to be used on pre-trained models due to the fact that they are not involved in the training process.
We focus on these types of detectors in this work.
These detectors use various methods; we provide a brief overview of some techniques in the related work section below, but refer to~\cite{yang_generalized_2024} for a more comprehensive discussion.

While the current state-of-the-art in OOD detection achieves good performance across a variety of measures, many crucially fall short in providing an explanation for why a particular input is rejected.
This hinders the ability to act on some detections in all scenarios where humans are in the loop to control the system~\cite{ruff_unifying_2021}. 
To alleviate this issue, we propose an OOD detector that utilises methods from the explainability domain to not only detect OOD points but also explain why they were flagged as OOD.
Our OOD detector uses counterfactual distance (the distance between an input and its counterfactuals) as an OOD score, simulating the distance from a point to decision boundaries.
Several works have shown a correlation between a point being OOD and being further from ID points and closer to decision boundaries~\cite{sun_out--distribution_2022,liu_fast_2024}.
As we report in this paper, the counterfactual distance is more accurate than the approximated distances used by other methods, e.g.,~\cite{liu_fast_2024}, and yields better ID-OOD separability.
Crucially, we can use the counterfactuals generated by our method to explain why the input was flagged as OOD without incurring any additional cost.
We present an illustration of our approach in Figure~\ref{fig:illustration} to help visualise our method.

There have been some works that tackle the explainability of OOD detectors by outputting heatmaps~\cite{liznerski_explainable_2021} or by generating counterfactual explanations after a point is classified as OOD~\cite{liznerski_reimagining_2024}.
However, by using counterfactual explanations to compute OOD detection scores, our method can both detect OOD points and explain them through counterfactuals without added overhead.

\paragraph{Contributions.} 
In summary, our contribution is as follows:
\begin{itemize}
\item We present a post-hoc counterfactual-distance-based OOD detection method, including both a method to compute these distances in input space and a method to reduce runtime costs by computing them in the embedding space. 
\item We show that we outperform state-of-the-art methods on a variety of widely used benchmarks, including CIFAR-10 and CIFAR-100~\cite{krizhevsky_learning_2009}, and ImageNet-200~\cite{zhang_openood_2024}, a subset of ImageNet-1K~\cite{deng_imagenet_2009}.
\item We propose an explanation framework for our OOD detector. We illustrate how, by using counterfactuals generated during the detection process, we can enhance the interpretability of OOD detection without incurring additional runtime costs, thereby yielding more directly interpretable results.
\end{itemize}

The remainder of this paper is organised as follows.
Below, we briefly discuss related work. Section~\ref{sec:background} introduces the necessary background on DNNs, OOD detection, and counterfactuals.
Section~\ref{sec:cf_dist} details our proposed counterfactual distance OOD score. Finally, Section~\ref{sec:eval} evaluates our method against standard baselines and presents our explanation framework alongside a case study.

\paragraph{Related Work.}
A primary approach in OOD detection involves defining a score function based on the outputs of a Deep Neural Network. 
For example, the Maximum Softmax Probability (MSP)~\cite{hendrycks_baseline_2017} uses a simple threshold on the highest softmax score, while ODIN~\cite{dcbe7abf4db64d1b89bf9802585660ed} enhances this by applying temperature scaling and input perturbations to increase the separation between in-distribution and OOD data. Energy-based scoring offers another popular alternative~\cite{lee_simple_2018}. A key limitation of these methods, however, is their vulnerability to model overconfidence, where a model can produce erroneously high-confidence predictions for OOD inputs~\cite{nguyen_deep_2015}.

To address this and produce more robust scores, a second line of work focuses on reducing noise by pruning or re-weighting neuron activations. ReAct~\cite{sun_react_2021} clips high-magnitude activations in the penultimate layer, observing that OOD inputs tend to produce activations with higher variance than their in-distribution counterparts. Going further, other methods assess the importance of different network components. DICE~\cite{sun_dice_2022} prunes noisy activations by ranking weights based on their contribution, while LINe~\cite{ahn_line_2023} employs Shapley values to identify the most salient neurons before applying activation clipping.

Distance-based methods have also emerged as top performers. These approaches score inputs based on their distance to the training data's distribution in some feature space. This includes using the k-th nearest neighbour distance in the model's embedding space as an OOD score~\cite{sun_out--distribution_2022}. The Fast Decision Boundary based Detector (fDBD)~\cite{liu_fast_2024} calculates an approximate distance to class decision boundaries, leveraging the insight that OOD points often reside closer to these boundaries than in-distribution points. 
Our work advances this concept by computing a more precise counterfactual distance to the decision boundary, thereby yielding a more discriminative OOD score.

The use of counterfactuals for the OOD detection task itself is a key novelty of our work. 
While counterfactual distance has been explored in adjacent domains, such as for membership inference attacks~\cite{pawelczyk_privacy_2023}, its role in OOD has been confined to post-hoc explanation. 
For instance, recent frameworks propose generating counterfactuals to interpret and explain \textit{why} a pre-existing detector flagged an input as an anomaly~\cite{liznerski_reimagining_2024}. While this aids interpretability, it does not contribute to the detection process itself. 
In contrast, our method is the first to integrate the counterfactual distance directly into the core scoring function. 
This holistic approach allows us to produce both a state-of-the-art OOD score and an explanation simultaneously, without incurring additional runtime costs.

\section{Background}\label{sec:background}

In this section, we define the notation to be used throughout this work and cover the necessary background for this paper.
We introduce DNN notation and then illustrate OOD detection.
Finally, we introduce counterfactual explanations.

\paragraph{Neural Networks.}
We consider a $d$-dimensional input space $\mathcal{X} \subseteq \mathds{R}^d$ and output space $\mathcal{Y} = \{1, 2, \dots, C\}$ of classes~\cite{goodfellow_deep_2016}.
Let $\mathcal{D}_{train}$ be a set of $m$ labelled training samples $\{(x^{(1)}, y^{(1)}, \dots, x^{(m)}, y^{(m)})\}$ where each $x^{(i)} \in \mathcal{X}$ and each $y^{(i)} \in \mathcal{Y}$.
Furthermore, let $f: \mathcal{X} \rightarrow \mathcal{Y}$ be a DNN trained on $\mathcal{D}_{train}$ and $f_{-1}(x)$ be the function returning the output of the penultimate layer of a DNN. 

\paragraph{OOD Detection.}
Once a DNN is deployed and used in the real world, it may encounter inputs unlike those it was trained and tested on. 
These types of inputs are considered OOD.
The goal of DNN OOD detection is to identify the outputs for these points as unsafe predictions, which may result in a warning system or require human intervention.
This is done by creating a score function $S: \mathcal{X} \rightarrow \mathds{R}$ whose output is used to determine whether an input is OOD or not:
    \begin{equation}
        \forall x \in \mathcal{X}, d(x) = 
        \begin{cases}
      \text{ID} & \text{if $S(x) \geq \tau$}\\
      \text{OOD} & \text{otherwise}
    \end{cases}
    \label{eq:perfect_ood}
    \end{equation}
where $d(x)$ is an OOD classifier and $\tau$ is a threshold.

\paragraph{Counterfactual Explanations.}
A counterfactual explanation shows how the input of a DNN can be minimally changed in order to alter its output~\cite{wachter_counterfactual_2017}.

\textbf{Definition 1 (Counterfactuals).}
\textit{
    Given an input $x \in \mathcal{X}$, a trained DNN $f$, and a distance function $dist : \mathds{R}^n \times \mathds{R}^n \rightarrow \mathds{R}^+$, $x' \in \mathcal{X}$ is a counterfactual for $x$ such that
    \begin{equation}
        \argmin_{x'} dist(x, x') \text{ subject to } f(x) \neq f(x').
    \end{equation}
}

Usually, the counterfactual is chosen such that $x'$ is also close to $x$ by some distance metric such as Manhattan or Euclidean distance (see, e.g., ~\cite{guidotti_counterfactual_2024} for a recent survey). 
Many methods of counterfactual search exist, and we will discuss some of these in Section~\ref{sec:cf_dist} and evaluate them within our framework in Section~\ref{sec:eval}.

\section{OOD Detection Using Counterfactual Distance}\label{sec:cf_dist}

In this section, we introduce our post-hoc counterfactual distance measure.
We first show how to calculate our score in input space and then introduce a variant where we calculate the distance in embedding space to reduce runtime costs.

Measures of distance have been used in the literature to inform whether an input is OOD or not.
In particular, some methods look at the distance between an input and the decision boundaries, postulating that DNNs are uncertain about points that reside close to these boundaries~\cite{liu_fast_2024}.
While previous methods use fast approximations to estimate boundary proximity, counterfactuals explicitly cross the decision boundary to a specific target class.
This provides a more precise, geometrically grounded distance metric that natively captures the minimal perturbation required to shift the model's prediction.
Our method reinforces the notion that ID features tend to be further away from the decision boundary than OOD features.
This method is post-hoc, meaning it is independent of the training procedure and can readily be used on any pre-trained DNN, regardless of its architecture.
Our method is based on the theoretical framework established by~\cite{liu_fast_2024}. For a comprehensive discussion of the underlying principles, we refer the reader to their original work.

\subsection{Counterfactual Distance Measure}
The first inclination is to compute counterfactual distances directly on the input space.
For an input $x$, we do this by taking the output of the pre-trained DNN, $f(x)$, and generate counterfactuals for each class in $\mathcal{Y}$ that is not equal to the output of the network on the input, $f(x)$.
The distances are computed using the $l_2$ norm and then we take the average and also regularise the scores following the methodology of~\cite{liu_fast_2024}. 
This regularisation is done by computing $\mu_{train}$, the mean of the input features of the training dataset, and computing the distance between $x$ and $\mu_{train}$.
Thus, our counterfactual distance measure is calculated as:
\begin{equation}
    \frac{1}{|\mathcal{Y}|-1} \sum_{y \in \mathcal{Y}, y \neq f(x)} \frac{\parallel CF(x, y) - x \parallel_2}{\parallel x - \mu_{train}\parallel_2}
    \label{eq:cf_dist_input}
\end{equation}
where $CF(x, y)$ is a function returning a counterfactual on an input $x$ for a target class $y$.

However, computing counterfactuals in the input space may be costly depending, e.g., on the number of input features and the size of the DNN.
We discuss how the runtime of the counterfactual search method is impacted by these parameters at the end of this section.
To reduce the runtime of our method, we also propose an alternative distance measure by moving to the embedding space.

\paragraph{Moving to Embedding Space.}
While we started our analysis on input space, we moved to embedding space to reduce the dimensionality of our calculations.
For each input $x$, we compute counterfactuals on $f_{-1}(x)$, the output of the penultimate layer of the DNN, instead of directly on $x$, which gives us a counterfactual in the embedding space.
In Equation~\ref{eq:cf_dist_emb}, we show how we take these counterfactual distances, similarly to what we did in input space.
\begin{equation}
    \frac{1}{|\mathcal{Y}|-1} \sum_{y \in \mathcal{Y}, y \neq f(x)} \frac{\parallel CF(f_{-1}(x), y) - f_{-1}(x) \parallel_2}{\parallel f_{-1}(x) - \mu_{train}\parallel_2}
    \label{eq:cf_dist_emb}
\end{equation}
where $CF(f_{-1}(x), y)$ is a function returning a counterfactual on $f_{-1}(x)$ with target class $y$.
When we move to the embedding space, $\mu_{train}$ is the mean over the training set in the embedding space, rather than in the input space.

We present the procedure for calculating the embedding-space OOD score in Algorithm~\ref{alg:cfdist2}.
The function GetCounterfactual takes in the embedding $z_x$ and a target class, $y$, from the known classes of the DNN excluding the predicted class for the input $x$.
We then use GetDistance to get the $l_2$ distance between $z_x$ and its counterfactual and take the average over the number of counterfactual distances we compute.
After we compute this distance, we normalise it with respect to the distance from the input to the mean of the training set in embedding space, which is precomputed.

\begin{algorithm}[tb]
    \caption{Counterfactual distance measure}
    \label{alg:cfdist2}
    \textbf{Input}: Sample input $x$\\
    \text{Known classes} $\mathcal{Y} = \{1, 2, \dots, C\}$ \\
    Pre-trained DNN $f$ \\
    \textbf{Output}: OOD score $D$
    \begin{algorithmic}[1] 
        \STATE $z_x \gets$ $f_{-1}(x)$
        \STATE $D \gets 0$
        \FOR{$y \in \{\mathcal{Y} \setminus f(x) \}$}
        \STATE $c_{x,y} \gets \textbf{GetCounterfactual}(z_x, y)$
        \STATE $D \gets D + \textbf{GetDistance}(z_x, c_{x,y})$
        \ENDFOR
        \STATE \textbf{return} $D$
    \end{algorithmic}
\end{algorithm}

\begin{table*}[!ht]
    \centering
    \resizebox{\textwidth}{!}{%
        \begin{tabular}{lcccccccccc}
            \toprule
            \multicolumn{1}{c}{} & \multicolumn{2}{c}{SVHN}& \multicolumn{2}{c}{iSUN}& \multicolumn{2}{c}{Textures}& \multicolumn{2}{c}{Places365}& \multicolumn{2}{c}{Average}\\
            Method & FPR95$\downarrow$ & AUROC$\uparrow$ & FPR95$\downarrow$ & AUROC$\uparrow$ & FPR95$\downarrow$ & AUROC$\uparrow$ & FPR95$\downarrow$ & AUROC$\uparrow$ & FPR95$\downarrow$ & AUROC$\uparrow$ \\
            \midrule
            MSP & 25.43 & 91.56 & 23.05 & 92.36 & 35.22 & 89.89 & 42.46 & 88.92 & 31.54 & 90.68\\
            ODIN & 68.54 & 84.70 & 27.20 & 94.63 & 67.89 & 86.94 & 70.40 & 85.08 & 58.51 & 87.84\\
            Energy & 34.20 & 91.98 & 27.69 &  93.24 & 52.01 & 89.46 & 54.69 & 89.25 & 42.15 & 90.98\\
            ViM & 19.01 & 94.58 & 21.12 & 94.20 & \textbf{21.14} & \textbf{95.16} & 41.44 & 89.50 & 25.68 & 93.36\\
            MDS & 25.88 & 91.17 & 32.09 & 90.11 & 28.03 & 92.69 & 47.69 & 84.91 & 33.42 & 89.72 \\
            KNN & 22.39 & 92.77 & 19.48 & 94.57 & 24.06 & 93.16 & 30.35 & 91.77 & 24.07 & 89.24\\
            DICE & 36.16 & 90.22 & 59.60 & 81.26 & 62.73 & 81.88 & 77.21 & 74.74 & 58.93 & 82.02\\
            fDBD & 22.50 & 92.93 & 19.24 & 94.57 & 24.35 & 93.13 & \textbf{29.16} & \textbf{92.01} & \textbf{23.81} & \textbf{93.16}\\
            \midrule
            Ours + NICE & \textbf{5.69} & \textbf{98.66} & 13.96 & \textbf{97.18} & 34.98 & 91.34 & 48.57 & 86.79 & 25.80 & 93.50\\
            Ours + NNCE & 5.79 & 98.55 & \textbf{13.69} & 97.01 & 37.83 & 88.99 & 60.59 & 80.49 & 29.47 & 91.26\\
            \bottomrule
        \end{tabular}%
    }
    \caption{Evaluation on CIFAR-10. We use the OpenOOD implementation for all benchmarks and average the values over three training runs from their saved models. We give results on our method using two different counterfactual search methods, NICE and NNCE. The best results are in bold.}
    \label{tab:cifar10}
\end{table*}

\begin{table*}[!ht]
    \centering
    \resizebox{\textwidth}{!}{%
        \begin{tabular}{lcccccccccc}
            \toprule
            \multicolumn{1}{c}{} & \multicolumn{2}{c}{SVHN}& \multicolumn{2}{c}{iSUN}& \multicolumn{2}{c}{Textures}& \multicolumn{2}{c}{Places365}& \multicolumn{2}{c}{Average}\\
            Method & FPR95$\downarrow$ & AUROC$\uparrow$ & FPR95$\downarrow$ & AUROC$\uparrow$ & FPR95$\downarrow$ & AUROC$\uparrow$ & FPR95$\downarrow$ & AUROC$\uparrow$ & FPR95$\downarrow$ & AUROC$\uparrow$ \\
            \midrule
            MSP & 58.43 & 78.68 & 51.05 & 82.06 & 61.79 & 77.32 & 56.65 & 79.21 & 56.98 & 79.32\\
            ODIN & 67.18 & 74.72 & 35.50 & 90.65 & 62.41 & 79.34 & 59.73 & 79.45 & 56.21 & 81.04\\
            Energy & 53.19 & 82.28 & 45.53 & 85.82 & 62.39 & 78.35 & 57.69 & 79.50 & 54.70 & 81.49\\
            ViM & 46.37 & 82.87 & 46.64 & 82.13 & 47.22 & 85.75 & 61.57 & 75.86 & 50.45 & 81.65\\
            MDS & 67.47 & 70.32 & 77.48 & 63.77 & 70.21 & 76.42 & 79.19 & 63.41 & 73.59 & 68.48\\
            KNN & 51.48 & 84.26 & 47.47 & 84.26 & 53.63 & 83.66 & 60.76 & 79.42 & 53.33 & 82.90\\
            DICE & 48.96 & 84.45 & 46.13 & 85.37 & 64.26 & 77.63 & 59.42 & 78.31 & 54.69 & 81.44\\
            fDBD & 53.53 & 80.67 & 41.36 & 86.10 & 53.60 & 81.18 & 57.18 & 79.85 & 51.42 & 81.95\\
            \midrule
            Ours + NICE & 0.40 & 99.86 & 3.40 & 99.31 & 34.67 & 93.17 & 50.75 & 88.78 & 22.30 & 95.28\\
            Ours + NNCE & \textbf{0.04} & \textbf{99.99} & \textbf{0.44} & \textbf{99.80} & \textbf{19.21} & \textbf{95.88} & \textbf{35.47} & \textbf{92.54} & \textbf{13.79} & \textbf{97.05}\\
            \bottomrule
        \end{tabular}%
    }
    \caption{Evaluation on CIFAR-100. We use the OpenOOD implementation for all benchmarks and average the values over three training runs from their saved models. We give results on our method using two counterfactual search methods, NICE and NNCE. The best results are in bold.}
    \label{tab:cifar100}
\end{table*}

On top of the reduction in runtime, we found that by taking counterfactual distance from a close to output layer of the DNN, we were able to better separate the ID and OOD data.
This finding is in line with other methods that use close-to-output-layer activations to improve OOD separability~\cite{henzinger_outside_2020,liu_fast_2024}.
Moving our counterfactual distance calculation to the embedding space creates a more accurate and efficient method than computing distances on the input space.

\paragraph{Remark.} 
The runtime of our approach is dependent on our choice of counterfactual explanation method, whether it is based on Mixed Integer Linear Programming (MILP)~\cite{mohammadi_scaling_2021}, gradient-based optimisations~\cite{wachter_counterfactual_2017}, or nearest neighbors~\cite{brughmans_nice_2024}.
For example, approaches that use MILP are NP-complete whereas nearest-neighbour based counterfactual search methods have a runtime that is quadratic in the number of features of the input~\cite{leofante_promoting_2024}.
The choice of counterfactual method is crucial to the efficiency of calculating our OOD score, as it dominates the runtime, both in input space and in embedding space.
In Section~\ref{sec:eval}, we will show how the choice of counterfactual method impacts the accuracy and runtime of our OOD detector.

\section{Evaluation}\label{sec:eval}
In the previous section, we described how to compute a counterfactual-distance-based OOD score and methods by which to reduce the runtime costs of our scoring system.
In this section, we evaluate our method using the embedding space calculations against several standard baselines on commonly used datasets for OOD detection.

We evaluate our method in the following experiments:
\begin{itemize}
    \item In Section~\ref{subsec:cifar_benchmarks} experiments were run with our embedding-space OOD score on both CIFAR-10 and CIFAR-100~\cite{krizhevsky_learning_2009}.
    We demonstrate the accuracy of our method against the state-of-the-art.
    \item In Section~\ref{subsec:imagenet200} we present results on the ImageNet-200 benchmark comparing against eight baseline methods and commonly used OOD datasets.
    \item  Finally, in Section~\ref{subsection:generatingcf}, we show how our counterfactuals can be used to generate explanations.
    We present an example of an explanation our method can produce with MNIST.
\end{itemize}

\subsection{Experimental Setup}\label{subsec:experimental_setup}

\paragraph{Datasets and Architectures.} 
To ensure a rigorous and direct comparison against the state of the art, our experimental protocol is aligned with the standards established by leading contemporary works in this area~\cite{zhang_openood_2024}.
We evaluate our method on CIFAR-10, CIFAR-100~\cite{krizhevsky_learning_2009}, and ImageNet-200~\cite{zhang_openood_2024}, using pre-trained ResNet-18 models~\cite{he_deep_2016} provided by OpenOOD~\cite{yang_openood_2022}. For each benchmark, we use four standard OOD datasets for evaluation. We also use MNIST~\cite{lecun_mnist_1998} for a case study with a custom 3-layer CNN. 
Full details on the datasets and DNNs can be found in the appendix.

\paragraph{Counterfactual and Baseline Methods.} 
Our detector is evaluated using two nearest-neighbour-based counterfactual explanation methods: NICE~\cite{brughmans_nice_2024} and NNCE~\cite{guidotti_counterfactual_2024}.
We focus on nearest neighbour methods due to their lower runtime costs compared to other methods such as MILP-based.
However, our detector can be easily extended to use any counterfactual explanation method.
Implementation details of these methods are included in the appendix.
We compare our method against eight state-of-the-art baselines, including MSP~\cite{hendrycks_baseline_2017}, ODIN~\cite{dcbe7abf4db64d1b89bf9802585660ed}, Energy~\cite{liu_energy-based_2020}, and fDBD~\cite{liu_fast_2024}. 
A complete list and description of baseline methods can be found in the appendix. 
All experiments were conducted on a machine with an AMD EPYC 7453 28-Core CPU and 512GB of RAM.

\paragraph{Evaluation Metrics.}
To assess the effectiveness of our method in distinguishing between ID and OOD samples, we report two standard metrics.
First, we use the False Positive Rate at 95\% True Positive Rate (FPR95).
This metric measures the percentage of OOD examples incorrectly accepted as ID (false positives) when the detection threshold is set to correctly classify 95\% of the ID data.
FPR95 is particularly relevant for safety-critical contexts, as it quantifies the risk of accepting OOD inputs at a high level of operational recall. 
Lower values indicate better performance.
Second, we report the Area Under the Receiver Operating Characteristic (AUROC).
This provides a threshold-independent measure of the detector's separability.
It can be interpreted as the probability that a randomly selected OOD example is assigned a higher anomaly score than a randomly selected ID example.
An AUROC of 50\% indicates random performance, while 100\% indicates perfect separation. 
Higher values are desirable.

\begin{table*}[!ht]
    \centering
    \resizebox{\textwidth}{!}{%
        \begin{tabular}{lcccccccccc}
            \toprule
            \multicolumn{1}{c}{} & \multicolumn{2}{c}{SSB-hard}& \multicolumn{2}{c}{iNaturalist}& \multicolumn{2}{c}{Textures}& \multicolumn{2}{c}{OpenImage-O}& \multicolumn{2}{c}{Average}\\
            Method & FPR95$\downarrow$ & AUROC$\uparrow$ & FPR95$\downarrow$ & AUROC$\uparrow$ & FPR95$\downarrow$ & AUROC$\uparrow$ & FPR95$\downarrow$ & AUROC$\uparrow$ & FPR95$\downarrow$ & AUROC$\uparrow$ \\
            \midrule
            MSP & \textbf{66.09} & 80.32 & 35.39 & 90.13 & 44.43 & 88.37 & 35.21 & 89.24 & 45.28 & 87.01\\
            ODIN & 73.52 & 77.17 & 22.49 & 94.34 & 42.96 & 90.67 & 37.29 & 90.10 & 44.07 & 88.07\\
            Energy & 69.97 & 79.72 & 26.41 & 92.52 & 41.29 & 90.80 & 36.80 & 89.20 & 43.62 & 88.06\\
            ViM & 71.59 & 74.00 & 26.81 & 91.38 & \textbf{20.19} & 94.77 & 33.96 & 88.21 & 38.14 & 87.09\\
            MDS & 83.60 & 58.52 & 58.13 & 75.34 & 58.13 & 79.25 & 67.87 & 70.15 & 66.93 & 70.81\\
            KNN & 73.90 & 76.88 & 24.43 & 93.97 & 24.52 & \textbf{95.30} & 32.94 & 90.17 & 38.95 & 89.08\\
            DICE & 71.02 & 78.95 & 29.74 & 91.77 & 40.79 & 91.54 & 38.82 & 89.03 & 45.09 & 87.82\\
            fDBD & 66.66 & 80.49 & \textbf{16.55} & 95.66 & 29.61 & 92.95 & 27.20 & 91.73 & 35.01 & 90.21\\
            \midrule
            Ours + NNCE & 73.61 & \textbf{80.60} & 23.55 & \textbf{96.32} & 35.48 & 93.65 & \textbf{1.55} & \textbf{99.63} & \textbf{33.55} & \textbf{92.55}\\
            \bottomrule
        \end{tabular}%
    }
    \caption{Evaluation on ImageNet-200. We use the OpenOOD implementation for all benchmarks and average the values over three training runs from their saved models. We give results on our method using the counterfactual search method NNCE. The best results are in bold.}
    \label{tab:imagenet200}
\end{table*}

\subsection{Evaluation on CIFAR Benchmarks}\label{subsec:cifar_benchmarks}

In this section, we provide experimental results on the accuracy of our method versus some commonly utilised benchmarks on both CIFAR-10 and CIFAR-100.
For both datasets, we average over three pre-trained ResNet-18~\cite{he_deep_2016} networks provided by OpenOOD~\cite{yang_openood_2022}.
We also utilise OpenOOD's API to obtain our results for the eight state-of-the-art methods we compare against for consistency.

\paragraph{CIFAR-10 Results.}
For our evaluation on CIFAR-10, we use the test set of 10,000 images for our ID samples. 
Results can be found in Table~\ref{tab:cifar10}, where we report results on the four OOD datasets and averages.
Our method outperforms all baselines against SVHN and iSUN and, on average, is in line with the results of fDBD, KNN, and ViM.

The results are generally stronger when comparing our method to the methods that score model uncertainty (MSP, ODIN, and Energy).
The FPR95 and AUROC for our method show more accuracy on the Textures dataset than ODIN and Energy and marginally outperform MSP.
On the Places365 dataset, MSP outperforms with a 42.46\% FPR95 versus our NICE method with 48.57\% FPR95.
However, on average, our method outperforms these methods.

Overall, on Places365 and Textures, we see that ViM, MDS, KNN, and fDBD outperform our method.
On Textures, our reported FPR95 is 34.98\% versus 21.14\% with ViM and on the Places365 dataset, our method produces a 48.57\% FPR95 while fDBD achieves 29.16\%.
On average, we are competitive across all methods, with fDBD, KNN, and ViM outperforming our method by about 1.99\%, 1.73\%, and 0.12\%, respectively.

Comparing NICE and NNCE, we see that NICE gave us slightly better results overall, but the NNCE method is comparable.
The most significant difference between the two is on the Places365 dataset, where we see a 12.02\% difference in FPR95.
While the results on CIFAR-10  are not consistently outperforming other methods, the results are best on SVHN using NICE with 5.69\% FPR95 and on iSUN using NNCE with 13.69\% FPR95, respectively.
To illustrate these differences, we look at the OOD scores produced by both methods to compare the differences in calculated distances.
The average OOD score using NICE on the ID data is 0.14, and on SVHN, it is 0.07.
For NNCE, these numbers are 0.32 for ID data and 0.19.
We see that NICE can find closer counterfactuals, as the distances are smaller than NNCE found.

\paragraph{CIFAR-100 Results.}\label{subsec:cifar100}
Our evaluation on CIFAR-100 shows promising results, increasing accuracy over all benchmarks on all OOD datasets. 
Our performance on CIFAR-100 stays consistent with our results from CIFAR-10, while other methods show a deterioration in accuracy.

In the previous set of experiments on CIFAR-10, we saw that ViM, KNN, MDS, and fDBD outperformed our method on some datasets.
However, when evaluating our method on CIFAR-100, the results for both NICE and NNCE improve upon FPR95 and AUROC against all baseline methods.
We hypothesize that this could be because computing distances to more decision boundaries can give more accuracy to our OOD score.
On SVHN, we see the best results with NNCE achieving a 0.04\% FPR95 and 99.99\% AUROC, where the next best result, with ViM, shows a 19.01\% FPR95 and a 94.58\% AUROC.
SVHN was similarly our best dataset on CIFAR-10.

Overall, looking at the average performance, our method achieves a 13.79\% FPR95 and 97.05\% AUROC, which is an improvement of 36.66\% FPR95 compared to the next best method, ViM which achieves an FPR95 of 50.45\%.
This is approximately a 72.67\% reduction in FPR.

The results on CIFAR-100 show the impact that a more accurate distance measure can have on the accuracy of the OOD score.
Compared to fDBD, we show an improvement of 37.63\% on FPR95 and 15.10\% on AUROC.

While NICE outperformed NNCE on the CIFAR-10 experiments, we see that NNCE shows stronger results on CIFAR-100.
We hypothesise that this outperformance is due to the structure of the underlying dataset, and these results show how accuracy is dependent on the choice of counterfactual search method.
In Section~\ref{subsec:imagenet200}, we provide experiments on ImageNet-200 to see how our method performs on a larger dataset.
We also include an ablation study on CIFAR-100 in the appendix to further investigate how we can reduce runtime costs.

\subsection{Evaluation on ImageNet-200}\label{subsec:imagenet200} 
In this section, we provide experimental results on ImageNet-200 with some commonly used OOD datasets in order to evaluate our method on a larger dataset.
For this benchmark, we utilise a pre-trained ResNet-18 network from OpenOOD~\cite{yang_openood_2022}.
We compare to the same eight baseline OOD detectors as we did in the previous section, also using the OpenOOD implementations of these methods.
Due to runtime constraints introduced by utilising a larger dataset, we only provide results using the counterfactual search method NNCE.

We present our results on the four OOD datasets and the average results in Table~\ref{tab:imagenet200}.
Our results show that our method outperforms the baselines on OpenImage-O with a 1.55\% FPR95 and 99.63\% AUROC, which is an improvement over the next best results with fDBD which obtained a 27.20\% FPR95 and a 91.73\% AUROC.
We also obtain the best AUROC on SSB-hard with 80.60\%, though our FPR95 is a bit higher than all methods except MDS and KNN.
Similarly, our method has the highest AUROC on iNaturalist with 96.32\%, though our FPR95 was a bit higher than ODIN and fDBD.
On Textures, the best FPR95 is found with ViM with a 20.19\% FPR95 and KNN has the best AUROC of 95.30\%.
Our results are comparable with a 35.48\% FPR94 and 93.65\% AUROC.
fDBD also has a better FPR95 than ours, but a lower AUROC.

On average, we outperform the baselines over these four OOD datasets with an overall FPR95 of 33.55\% and AUROC of 92.55\%.
These results show our method can separate ID and OOD data on larger and more complex datasets.
As in our experiments on the CIFAR benchmarks, these results show that the structure of the underlying datasets introduces some variance in performance.
While some methods outperform ours on certain OOD datasets, our results are obtained via a counterfactual search method that finds an approximate counterfactual via nearest unlike neighbours, rather than a more optimised search.
Thus, the distance we calculate to the decision boundary will be coarser.
The results could be further improved with a more precise counterfactual search, but this would increase runtime.

Furthermore, to address runtime costs associated with scaling, we conducted an ablation study on CIFAR-100 (detailed in Appendix~\ref{subsec:ablation}). Computing counterfactuals for 100 classes takes 8.74 ms per instance. Reducing the evaluated target classes ($k$) to 50 reduces this runtime by 4.21 ms per instance, incurring a negligible 0.27\% penalty to FPR95.
While baseline approximations like fDBD operate faster (roughly 0.53 ms), our framework can be dynamically tuned to balance accuracy and latency, framing this gap as the computational trade-off required for simultaneous detection and zero-overhead interpretability.

\subsection{Generating Counterfactual Explanations}\label{subsection:generatingcf}
Beyond accurately distinguishing between ID and OOD data, our framework serves as an interpretability tool with negligible computational overhead.
By analysing the embedding space distances computed during detection, we can extract~\textit{exemplar-based explanations} highlighting which features of an OOD input diverge from the DNN's learned representations.

While generative approaches (e.g., using autoencoders) can visualise counterfactuals, they often require training auxiliary models and will increase runtime costs. 
Instead, we leverage the pre-calculated distances to identify specific training samples that define the local decision boundary. 
We map the embeddings of the training set back to their original input images to retrieve two sets of neighbours for a given input $x$:
\begin{enumerate}
    \item \textbf{Nearest Like Neighbours (NLNs):} The $k$ closest training samples belonging to the class predicted by the model. These show what the model \textit{thinks} the input looks like.
    \item \textbf{Nearest Unlike Neighbours (NUNs):} The $k$ closest training samples from other classes. These act as counterfactuals, answering the question: \textit{What is the closest known concept to this unknown input?}
\end{enumerate}

\begin{figure}[!ht]
\centering
\includegraphics[width=0.9\linewidth]{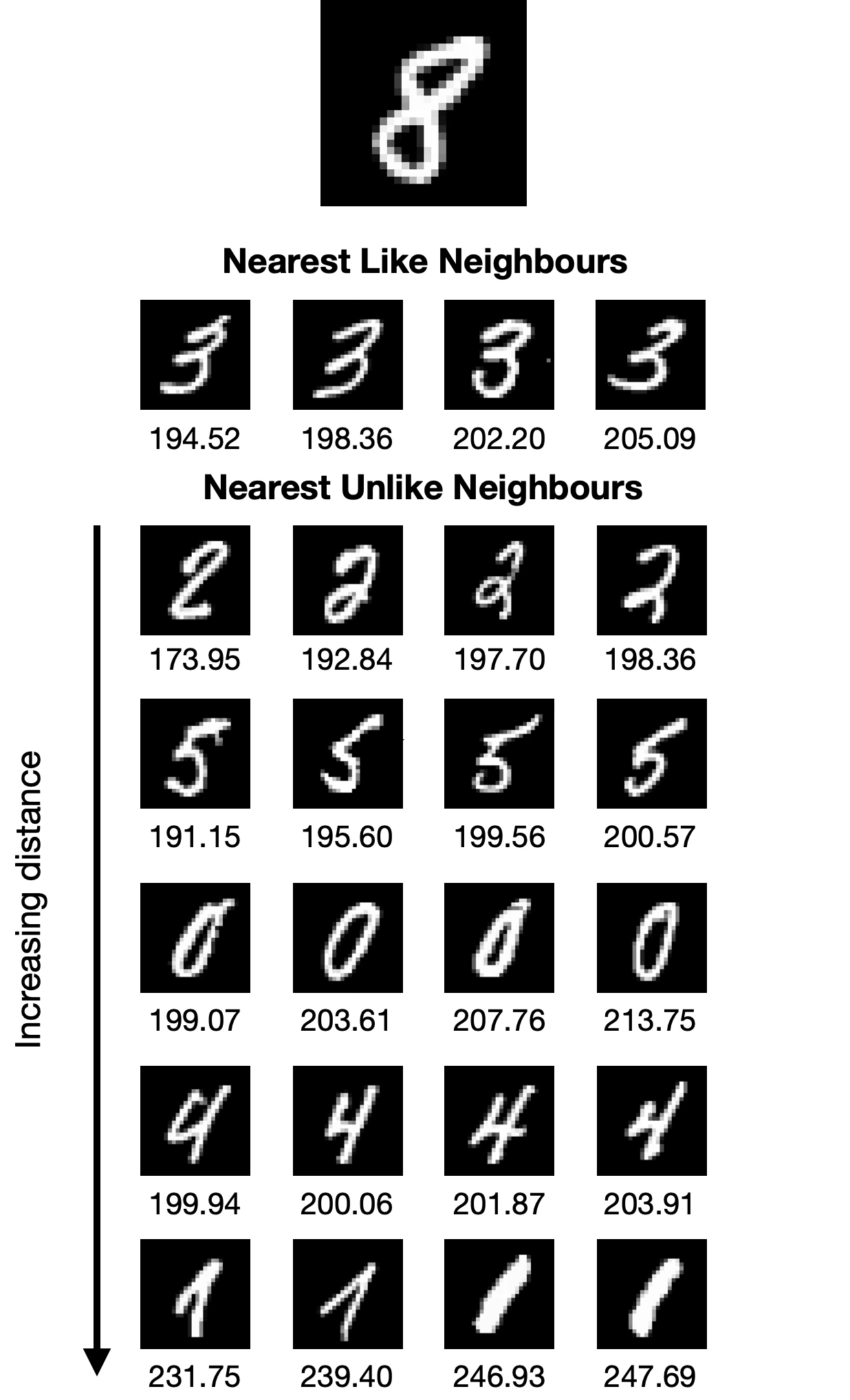}
\caption{Exemplar-based explanation for a novel class. The input (top) is an `8' (OOD), misclassified as a `3'. The top row shows the Nearest Like Neighbours (closest 3's). The bottom rows show Nearest Unlike Neighbours from other classes. Distances reveal the input is actually closer to class `2' than the predicted class, highlighting the model's uncertainty.}
\label{fig:counterfactual_example}
\end{figure}

This comparative visualisation helps users discern whether a sample is OOD due to semantic novelty (a new class) or covariate shift (a known class with unusual features). We detail this process in Algorithm~\ref{alg:cf_explainer}.

\begin{algorithm}[tb]
    \caption{Exemplar-Based Explanation Generation}
    \label{alg:cf_explainer}
    \small
    \textbf{Input}: Input $x$, Predicted Class $\hat{y} = f(x)$, Training Set $\mathcal{D}_{train}$, Neighbours $k$\\
    \textbf{Output}: Explanation Set $E$
    \begin{algorithmic}[1]
        \STATE $z_x \gets f_{-1}(x)$
        \STATE $E_{like} \gets \text{GetkNN}(z_x, \mathcal{D}_{train}^{\hat{y}}, k)$ \COMMENT{Retrieve NLNs}
        \STATE $E_{unlike} \gets \emptyset$
        \FOR{$c \in \text{Classes} \setminus \{\hat{y}\}$}
            \STATE $S_c \gets \text{GetkNN}(z_x, \mathcal{D}_{train}^{c}, k)$ \COMMENT{Retrieve NUNs per class}
            \STATE $E_{unlike} \gets E_{unlike} \cup S_c$
        \ENDFOR
        \STATE \textbf{return} $\{E_{like}, \text{SortByDistance}(E_{unlike})\}$
    \end{algorithmic}
\end{algorithm}

\paragraph{Case Study: MNIST.}
To demonstrate the utility of these explanations, we train a CNN on MNIST classes 0--5 (ID), treating 6--9 as OOD. This split allows us to evaluate detection on unseen classes, achieving an FPR95 of 44.26\% and an AUROC of 89.20\%.

Figure~\ref{fig:counterfactual_example} illustrates a novelty detection scenario where an OOD input (an `8') is incorrectly predicted as a `3'. The explanation first displays the NLNs (top row), showing the 3's that most structurally resemble the input; however, the visual discrepancy remains high. Crucially, the NUNs (bottom rows) reveal that the input is actually closer in the embedding space to the `2' class than to its predicted `3' class. By presenting these conflicting nearest neighbours---where the input fails to strongly align with any retrieved images---the user can visually confirm that the input represents a novel concept unsupported by the training distribution. (A second case study demonstrating distribution shift is provided in Appendix~\ref{subsec:case_study}.)

\paragraph{Summary.}
By providing both factual (NLNs) and counterfactual (NUNs) evidence, our method reduces over-reliance on the model's prediction. It allows users to distinguish between inputs that are truly novel (unlike anything seen before) and those that are merely confusing boundary cases (visually similar to a non-predicted class).

Overall, our counterfactual-distance-based scoring method is on par with or outperforms the state-of-the-art in OOD detection.
It achieved promising results, specifically on CIFAR-100.
However, our method stands out from the current baselines by providing interpretability as a by-product.
We can produce counterfactual and factual explanations to aid end users in understanding why an input is OOD.

\section{Discussion and Conclusion}\label{sec:conclusion}

In this paper, we introduced a novel method to perform OOD detection using counterfactual distance. We showed two variations of this method: one operating in input space and a more lightweight approach using the outputs of the penultimate layer of the network. 
Finally, we presented a framework to describe how we can easily generate explanations without added runtime costs to aid in determining why an input is classed as OOD. Our method outperforms the state-of-the-art on CIFAR and ImageNet-200 benchmarks, and we show methods to reduce the runtime costs of our method.
Crucially, because the OOD detection score is derived directly from the counterfactual distance, this interpretability is baked into the model, incurring zero marginal runtime cost.

This paper opens interesting avenues for future work, notably investigating lightweight, GPU-optimised counterfactual search methods to improve runtime assurance on larger datasets and complex DNN architectures.
It would also be valuable to investigate how utilising different or robust counterfactual search methods impacts OOD detection accuracy.

A limitation of the current work is that the explanation framework is demonstrated through qualitative case studies, making its practical utility difficult to assess quantitatively.
Future work will focus on conducting comprehensive user studies and developing quantitative evaluation metrics to ascertain the value of these explanations in practical deployment.

\section*{Acknowledgements}
Maria Stoica is supported by the UKRI Centre for Doctoral Training in Safe and Trusted Artificial Intelligence (EP/S023356/1).
Alessio Lomuscio is supported by a Royal Academy of Engineering Chair in Emerging Technologies.

\bibliographystyle{named}
\bibliography{references}

\clearpage
\appendix

\section{Appendix}\label{sec:appendix}

\subsection{Counterfactual Explanation via Nearest Unlike Neighbours}
The Nearest-Neighbour Counterfactual Explainer (NNCE) method for generating counterfactual explanations is to identify the \emph{nearest unlike neighbour} of a given instance \( x \): the closest point \( x_{nn} \) in the training set that receives a different prediction under the target classifier~\cite{guidotti_counterfactual_2024}. 
The counterfactual explanation is \( x_{nnce} \), the instance with the smallest such distance that leads to a different prediction.

To improve upon this baseline, the NICE algorithm (Nearest Instance Counterfactual Explanations) extends this approach by iteratively constructing counterfactuals through feature substitutions from \( x_{nnce} \). Starting from \( x \), NICE explores combinations of feature changes using a depth-first search, replacing values with those from \( x_{nnce} \) to generate valid counterfactuals. 
Several variants of NICE allow for user-defined preferences: minimising the number of changed features (sparsity), minimising the total distance (proximity), or maximising data manifold plausibility. 
Unlike gradient-based methods, NICE is model-agnostic, applicable to any classifier, and guarantees explanation coverage.

\subsection{Experimental Setup}

\paragraph{Datasets.}
We present two experiments for the CIFAR benchmarks, one on CIFAR-10 and another on CIFAR-100~\cite{krizhevsky_learning_2009}. 
Both of these experiments use the following four OOD datasets, following the methodology from fDBD: SVHN~\cite{netzer_reading_2011}, iSUN~\cite{xu_turkergaze_2015}, Places365~\cite{zhou_places_2018}, and Textures~\cite{cimpoi_describing_2014}.

In Section~\ref{subsec:imagenet200}, we present experiments for ImageNet-200~\cite{zhang_openood_2024}, which is a 200 class subset of the ImageNet-1K dataset~\cite{deng_imagenet_2009}.
We use the following four OOD datasets:
SSB-hard~\cite{vaze2022openset},
iNaturalist~\cite{van_horn_inaturalist_2018},
Textures~\cite{cimpoi_describing_2014},
and OpenImage-O~\cite{wang_vim_2022}.

In Section~\ref{subsection:generatingcf}, we present a case study on generating explanations on the MNIST dataset~\cite{lecun_mnist_1998}.
In order to create OOD points, we split the MNIST data by taking the first six classes (0-5) as ID inputs and the rest (6-9) as OOD.

\paragraph{Architectures.}
The experiments on the CIFAR datasets and ImageNet-200 are run on pre-trained ResNet-18~\cite{he_deep_2016} models.
We average our results over three separate training runs provided by OpenOOD~\cite{yang_openood_2022}.
For MNIST, we train a three layer convolutional neural network (CNN) on the first six classes as described above.
The CNN achieves 98\% accuracy on the ID test set.

\paragraph{Baseline Methods.}
We compare our method against eight baseline methods.
The first three methods compute OOD scores directly on the output of the DNN.
MSP~\cite{hendrycks_baseline_2017} calculates the maximum softmax probability of all inputs as an OOD score, ODIN~\cite{dcbe7abf4db64d1b89bf9802585660ed} extends MSP through temperature scaling and adversarial perturbations, and Energy~\cite{liu_energy-based_2020} calculates an energy score on the outputs of the DNN.
A few methods utilise distance metrics, such as MDS~\cite{lee_simple_2018}, which estimates a Gaussian distribution over the latent space and uses the Mahalanobis distance for their score, KNN~\cite{sun_out--distribution_2022}, which computes the distance to the kth nearest neighbour and fDBD~\cite{liu_fast_2024} which calculates the average distance to decision boundaries.
We also include two other methods utilising different approaches:
ViM~\cite{wang_vim_2022} which calculates an OOD score based on the feature space and the logit space and DICE~\cite{sun_dice_2022} which takes advantage of sparsification to only consider important weights to obtain an energy-based OOD score.

\subsection{Runtime Analysis on CIFAR-100}\label{subsec:ablation}
Our experiments on CIFAR-100 show favourable results, outperforming all baselines.
We perform a runtime analysis of our method using NNCE on CIFAR-100 to further show the practical applicability of our approach.
We consider a variation of our method whereby instead of computing counterfactual distances for all possible target classes, we compute only $k$ counterfactuals.
We choose the top $k$ classes (excluding the predicted class) based on the output probability distribution from the DNN and compute the counterfactual distance to each of these classes.
The choice of which counterfactuals to compute can instead be determined by the user, for example if some classes are deemed more important based on some domain knowledge.

In Table~\ref{tab:cifar100_ablation}, we show results with our NNCE method on CIFAR-100 when computing counterfactuals for the top $k$ classes for $k=$ 25, 50, 75, and 100, the latter of which is equivalent to what we did in the previous section.
We also include the average number of milliseconds our scoring method takes, averaged over 100 inputs when run on a 28-core CPU with 512 RAM.

\begin{table}[tb]
    \centering
    \small
    \begin{tabular}{lccc}
        \toprule
        $k$  &  FPR95$\downarrow$ & AUROC$\uparrow$ & Avg Time (ms)\\
        \midrule
        25      & 14.93 & 96.85 & 2.24\\
        50      & 14.06 & 97.01 & 4.53\\
        75      & 14.10 & 96.95 & 6.65\\
        100     & 13.79 & 97.05 & 8.74\\

        \bottomrule
    \end{tabular}
    \caption{Ablation study on CIFAR-100 OOD benchmarks using our method with NNCE averaged over OOD datasets SVHN, iSUN, Textures, and Places365 and averaged over three training runs, using the same ResNet-18 network as in Table~\ref{tab:cifar100}. Here, $k$ is the number of classes we compute counterfactuals for and thus, how many decision boundaries we consider in our distance metric. Avg Time (ms) refers to the milliseconds it takes to compute our OOD score on a machine equipped with a 28-core CPU with 512GB RAM. We average the computation time over 100 inputs.}
    \label{tab:cifar100_ablation}
\end{table}

In Table~\ref{tab:cifar100_ablation}, we see that the FPR95 and the AUROC generally worsen as we reduce $k$, but interestingly, $k=50$ slightly outperforms $k=75$.
This suggests that the 25 extra classes we consider for $k=75$ do not provide us with more information, and they may add a bit of noise to our scoring method.

Even with only 25 classes, our average result beats all the baselines from Table~\ref{tab:cifar100}.
In this case, an end user could determine the right trade-off between accuracy and runtime costs in determining the $k$ that is best for a particular use case.
We also hypothesise that this choice is highly dependent on the datasets used, and with some domain knowledge, a user can select which $k$ classes to use in the distance measure.
For example, on CIFAR-100, we may decide that reducing $k$ to 50 is worth the 0.27\% worsening of the FPR95 if we also get a 4.21 millisecond runtime benefit.
This ablation study provides insight into utilising this scoring method on larger datasets with more complex classes.

Comparing our runtime to other methods, fBDB shows that their method can produce scores of 0.53 ms on CIFAR-10. 
Their experiments were run on a machine equipped with a Tesla T4 GPU.
We do not expect our method to be faster than theirs. 
However, exploring more lightweight methods of implementing the counterfactual search and utilising GPUs would mitigate some of this runtime difference.
To our knowledge, no counterfactual search methods are optimised for GPU hardware.

\subsection{Case Study: Distribution Shift}\label{subsec:case_study}
We next consider an input `5'. While `5' is a known class, this specific image is visually atypical, leading the model to misclassify it as a `0'.
The explanation framework retrieves the NLNs (class `0'), which look significantly different from the input. 
However, the NUNs reveal that the closest class in the embedding space is actually the correct class, `5'.
Unlike the previous novelty example, where the input looked unlike \textit{any} neighbour, here the input clearly aligns visually with the retrieved `5's. This distinction allows a human auditor to diagnose the error: the detector correctly flagged the sample as OOD because it lies in a sparse region of the manifold, but the explanation helps recover the true label.

\begin{figure}[tb]
    \centering
    \includegraphics[width=0.9\linewidth]{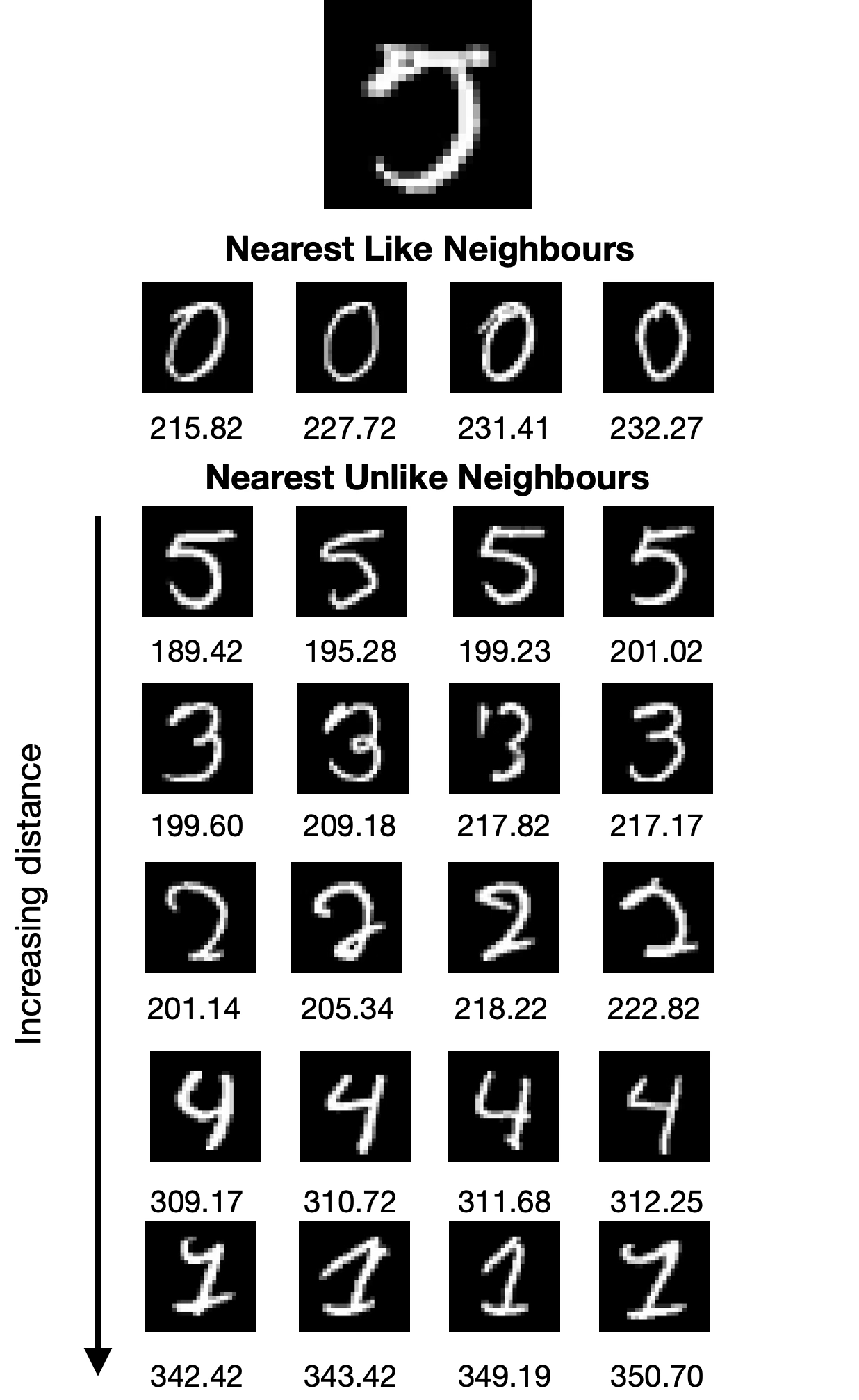}
    \caption{Explanation for a distribution shift. The input (top) is a `5' misclassified as a `0'. Despite the prediction, the explanation indicates that the input is semantically closer to the `5' class (Nearest Unlike Neighbours) than to the predicted `0' class (Nearest Like Neighbours), thereby aiding error diagnosis.}
    \label{fig:counterfactual_example_distribution_shift}
\end{figure}

\end{document}